\documentclass{article}

\usepackage[final,nonatbib]{nips_2016}

\usepackage[utf8]{inputenc} 
\usepackage[T1]{fontenc}    
\usepackage{url}            
\usepackage{booktabs}       
\usepackage{amsfonts}       
\usepackage{nicefrac}       
\usepackage{microtype}      

\title{Virtual Embodiment: A Scalable Long-Term Strategy for Artificial Intelligence Research}

\author{
 Douwe Kiela\\
 Facebook AI Research\\
 New York, NY 10017, USA\\
 \texttt{dkiela@fb.com} \\
\And
Luana Bulat, Anita L.~Ver\H{o}, Stephen Clark \\
Computer Laboratory, University of Cambridge\\
Cambridge CB3 0FD, UK\\
\texttt{\{ltf24,alv34,sc609\}@cam.ac.uk}\\
}

\begin{document}

\maketitle


\section{Introduction}

Meaning has been called the ``holy grail'' of a variety of scientific disciplines, ranging from linguistics to philosophy, psychology and the neurosciences \cite{Jackendoff:2002}. The field of Artifical Intelligence (AI) is very much a part of that list: the development of sophisticated natural language semantics is a \emph{sine qua non} for achieving a level of intelligence comparable to humans. Embodiment theories in cognitive science hold that human semantic representation depends on sensori-motor experience \cite{Barsalou:2008arp}; the abundant evidence that human meaning representation is grounded in the perception of physical reality leads to the conclusion that meaning must depend on a fusion of multiple (perceptual) modalities \cite{Meteyard:2008book}. Despite this, AI research in general, and its subdisciplines such as computational linguistics and computer vision in particular, have focused primarily on tasks that involve a single modality. Here, we propose \emph{virtual embodiment} as an alternative, long-term strategy for AI research that is multi-modal in nature and that allows for the kind of scalability required to develop the field coherently and incrementally, in an ethically responsible fashion.

Embodiment theory implies that the best way for acquiring human-level semantics is to have machines learn through (physical) experience: if we want to teach a system the true meaning of ``bumping into a wall'', we simply have to have it bump into walls repeatedly. Although this scenario shares similarities with human language acquisition, it is not (yet) a viable route: our current machine learning paradigms do not allow for the required rate of learning to make such a scenario feasible. With modern day state-of-the-art deep learning systems requiring millions of samples to solve highly specific tasks that are trivial to humans, it is reasonable to speculate that it would take much longer than a human lifespan for a \emph{physically} embodied agent to develop extensive linguistic capabilities, with current technology. We conjecture that such limitations apply to a much lesser extent to an agent that is \emph{virtually} embodied.

By virtual embodiment we mean to say that agents may collectively or individually acquire semantics by being embodied in a virtual, rather than a physical, world. Concretely, rather than having a physical robot learn to understand the world by physically bumping into physical walls, we would have virtual agents bump into virtual walls in a virtual world. Such virtual embodiment offers several key advantages:

\begin{enumerate}
\item Scalability and incremental development: The complexity of virtual worlds can develop in conjunction, i.e., scale, with the capabilities of artificial agents. This allows for a stepwise development towards general machine intelligence, rather than aiming for the end-goal without a concrete understanding of the challenges or consequences we will face when attempting to reach that end-goal.
\item Long-term feasibility: The performance ceiling of any agent is a function of the complexity of the virtual environment. Virtual worlds may initially not be overly complex, but they can grow in complexity as technology develops. This allows for a focused long-term research strategy that is feasible now, but will remain challenging in years and decades to come.
\item Rapid iteration: The fact that artificial agents are constrained by arbitrary parameters means that development can happen rapidly and iteratively, through agents learning from interacting both with humans and with each other. Rather than the extremes of either having one system solve small uni-modal tasks, or instead trying to solve the whole problem in a single attempt, we can improve iteratively, in an agile fashion, at great speed.
\item No requirement for continuous human involvement: Although interaction is necessary for embodied learning, virtual interactions need not require human involvement at each step, but may rather happen between agents themselves. This unburdens humans by foregoing the need for a constant supervised signal, as is currently often seen in machine learning applications, which also facilitates rapid development.
\item Ethical testability: Importantly, since artificial agents are exposed to a constrained environment, virtual worlds provide the ultimate testing ground for carefully fleshing out important ethical considerations in relation to artificial intelligence \cite{Bostrom:2003paperclip}, without any potentially damaging immediate consequences in the physical world. 
\end{enumerate}

For these reasons, we propose virtual embodiment as one of the best and most feasible strategies for instigating a stepwise development towards artificial general intelligence. In particular, we advocate the development and use of ``video games with a purpose'' to facilitate virtual embodiment. In what follows, we briefly outline some of the background that led to this proposal, explain why video games are suitable for the current purposes and list the desiderata for virtual embodiment-compatible video games to facilitate research in artificial intelligence.

\section{Grounding Semantics in Virtual Perception}

A fundamental problem of semantics is the grounding problem \cite{Harnad:1990}, which concerns the circularity in defining the meaning of a symbol through other symbols. In the context of Searle's famous Chinese Room argument \cite{Searle:1980bbs}, it can be phrased as: is it possible to learn Chinese from nothing but a (very sophisticated) Chinese dictionary? Modern representation learning approaches, including the word embeddings that have become popular in natural language processing, are exponents of the distributional hypothesis \cite{Harris:1954word}, which stipulates that you ``shall know the meaning of a word through the company that it keeps'' \cite{Firth:1957book}. In other words, semantic representation learning defines symbols through other symbols, which exposes it to the grounding problem. In contrast, there is abundant evidence that human meaning representation is grounded in physical reality and sensorimotor experience \cite{Jones:1991cd,Barsalou:1999bbs,Glenberg:2002pbr,Louwerse:2011tcs,Barsalou:2005book}.

Motivated by these theoretical considerations, the field of multi-modal semantics aims to ground semantic representations by introducing extra-linguistic, perceptual input into semantic models. Multi-modal semantic models lead to practical improvements in a variety of natural language processing tasks, ranging from resolving linguistic ambiguity \cite{Berzak:2015emnlp} to metaphor detection \cite{Shutova:2016naacl}. Beyond vision, there has also been work aimed towards auditory \cite{Lopopolo:2015iwcs,Kiela:2015emnlpa} and even olfactory \cite{Kiela:2015acl} grounding. However, most current multi-modal semantic models suffer from two important limitations. First, images and to a lesser extent sound files lack the element of \emph{time}, whereas temporal and sequential input are central aspects of language understanding. Second, these approaches lack any \emph{interaction}, which plays an important role in language acquisition: children learn basic language understanding by interacting with the environment, and build more intricate ``reflective reasoning'' on top of that foundation \cite{Landau:1998tcs}.



There has been work in linguistic grounding that allows for temporal aspects, for instance in videos \cite{Gupta:2009cvpr,Regneri2013:tacl,Yu:2015jair}, and both time and interaction, notably in the field of robotics \cite{Fitzpatrick:2003rsa,Coradeschi:2013ki,Bisk:2016naacl}. However, robotics does not currently constitute a suitable platform for language learning, since physical embodiment is not yet feasible. Virtual embodiment does not suffer from the same limitations. There has been recent work on grounding in virtual worlds, notably in video games \cite{Narasimhan:2015emnlp}. Work applying deep reinforcement learning to video games points the way towards agents learning from each other \cite{Mnih:2015nature,Silver:2016nature,Lazaridou:2016arxiv}. An alternative would be virtual or augmented reality, which offers the benefit of joint multi-modal data over time, but this crucially lacks the element of interaction.

Our position is very much aligned with recent proposals for new directions in AI research \cite{Mikolov:2015arxiv,Sukhbaatar:2015arxiv,Weston:2015arxiv,Johnson:2016malmo}. The particular problem of language features in these proposals to a varying extent, but we take it to be a core piece of any path toward artificial general intelligence, in line with recent attempts to make machines genuinely understand human language \cite{Hermann:2015nips,Wang:2016acl}. We specifically advocate multi-agent video games ``with a purpose'' \cite{VonAhn:2004chi}, rather than alternative virtual worlds that lack gamification, since they provide interesting platforms for humans to engage with for extended periods of time, without the explicit purpose of teaching machines to achieve a certain task.

\section{Desiderata}

It is worthwhile outlining the properties that video games might have if they are to be suitable platforms for developing AI through virtual embodiment. For that purpose, we propose a hierarchy\footnote{Inspired by the Kardashev scale for the sophistication of civilizations \cite{Kardashev:1964hierarchy}.} of the types of embodied manifestations an agent might have in a world. The same type hierarchy applies both to physical and to virtual worlds:

\begin{itemize}
\item Type 0: Agents perform basic first-order interactions with the world, with full or limited access to the objective world state. No intra-agent communication is required.
\item Type 1: As above, but without any state access. Communication may be used for sharing knowledge about the state of the world.
\item Type 2: As above, but with higher-order interactions, i.e., with an element of planning, strategy and non-monotonic reasoning. Communication is essential for sharing knowledge about the world.
\item Type 3: The world should be strictly non-deterministic and multi-modal. This makes communication essential for not only sharing knowledge about the world, but also for sharing plans and strategies.
\item Type 4: Agents should be multi-objective, that is, an agent's objective or reward function should be a weighted function of various objectives or rewards, that depend both on the state of the world and current plans and strategy.
\item Type 5: Multi-objective agents interact with and communicate about a non-deterministic world in such a way that it allows for them to plan ahead and form and execute sophisticated strategies.
\end{itemize}

The final type of embodiment corresponds to what biological agents are capable of performing in the physical world. It is much too large a leap for current technologies to achieve, but the benefit of virtual embodiment is that we can grow the complexity of the world together with the sophistication of artificial agents, which makes virtual embodiment suitable for being AI's next frontier. The real world is enormously complex, and performing common sense reasoning in such a complicated environment has long been one of AI's classic problems in the shape of the frame problem \cite{McCarthy:1969rai}. The frame problem is a function of the world's complexity, which makes it more manageable for virtual embodiment.

Most recent work has not extended beyond Type 1 embodiment, which means that the field has a long way to go. Specifically in the context of video games, we believe that development can proceed more rapidly if they are of mixed agency, meaning that both humans and artificial systems control agents in the virtual world; and carefully designed as a level playing field with a human bias, such that human agents have a slight upper hand, which means that e.g. the superior memory of machines should not affect in-game performance and that machines can learn from humans. To our knowledge no video game currently exists that satisfies these properties and which facilitates Type 5 embodiment.

\section{Conclusion}

We propose virtual embodiment, through video games, as a scalable long-term strategy for artificial intelligence research. Embodiment is essential for developing human-level natural language semantics, which we take to be a core aspect of artificial intelligence. Virtual embodiment allows for growing the complexity of virtual worlds in line with the sophistication of artificial agents, which makes it a suitable testing ground for artificial intelligence, in an ethically responsible manner.

\subsubsection*{Acknowledgments}

This research was enabled by the European Research Council PoC grant {\sc GroundForce}.

\bibliographystyle{unsrt}
\bibliography{references2}

\end{document}